\documentclass{Odyssey2026}

\interspeechcameraready

\title{Dataset Collection and Fine-Tuning of NLLB for German--English Translation}

\author[affiliation={1}]{Deeraj}{S. K.}
\author[affiliation={1}]{Yash}{Ingle}
\author[affiliation={1}]{Pruthwik}{Mishra}

\affiliation{Department of Artificial Intelligence}{Sardar Vallabhbhai National Institute of Technology}{India}
\email{u23ai050@coed.svnit.ac.in, u23ai062@coed.svnit.ac.in, pruthwikmishra@aid.svnit.ac.in}

\usepackage{comment}
\usepackage{soul}
\usepackage{xurl}
\urldef{\myhfurl}\url{https://huggingface.co/drjk16/NLLB-200-Distilled-600M\_finetuned\_Paragraph-Alligned}
\urldef{\mygiturl}\url{https://github.com/DeerajKarthik/paragraph\_aligned\_fine-tuning\_NLLB}

\begin{document}

\maketitle

\begin{abstract}
  This paper describes the curation of a high-quality parallel corpus from OPUS Books and the fine-tuning of Meta's NLLB-200 models for German--English translation. We detail a dataset extraction pipeline that filters 45k raw OPUS sentence pairs to 25k high-quality examples through encoding validation, length filtering, language identification, deduplication, and quality scoring. We present a reproducible fine-tuning methodology optimized for consumer GPUs using mixed precision, gradient accumulation, and gradient checkpointing. Our fine-tuned NLLB-600M model achieves statistically significant improvements (+3.4 BLEU, +2.6 chrF++, -3.2 TER) over the base model on literary and technical German-English translation. We demonstrate practical deployment by integrating the fine-tuned model into LinguaHaru, an existing open-source document translation system, achieving format-preserving translation across PDF and Office documents. All dataset preparation scripts, training code, and evaluation tools are provided for reproducibility at GitHub~\footnote{\mygiturl}. The fine-tuned model is available at HuggingFace~\footnote{\myhfurl}.
\end{abstract}

\section{Introduction}

Neural machine translation (NMT) has achieved remarkable progress through large-scale multilingual models like NLLB-200 \cite{NLLB2022}, M2M-100 \cite{Fan2021}, and mBART \cite{Liu2020}. However, these general-purpose models often underperform on specific domains or language pairs due to limited exposure to domain-specific vocabulary, style, and discourse patterns during pre-training. Fine-tuning on targeted parallel corpora has emerged as an effective approach to domain adaptation \cite{Chu2017,Farajian2017}.

This paper addresses the practical challenge of fine-tuning NLLB models for German--English literary and technical translation using freely available OPUS parallel corpora \cite{OPUS2016}. We focus on three key aspects: (1) systematic dataset curation from noisy web-scale corpora, (2) memory-efficient fine-tuning on consumer hardware, and (3) practical deployment in a document translation pipeline.

\subsection{Key Contributions}

Our work makes the following contributions:

\begin{itemize}
\item A systematic dataset extraction methodology for OPUS Books, applying multi-stage filtering to produce a curated 25k German--English parallel corpus from 45k raw sentence pairs
\item A reproducible fine-tuning pipeline for NLLB-600M optimized for 4GB consumer GPUs through mixed-precision training, gradient accumulation, and checkpointing
\item Quantitative evaluation with statistical significance testing, demonstrating +3.4 BLEU improvement over the base model
\item Practical integration into LinguaHaru \cite{LinguaHaruGitHub}, an open-source document translation system, enabling format-preserving translation across multiple document types
\item Complete open-source release of dataset preparation, training, and evaluation scripts with pinned dependencies
\end{itemize}

\subsection{Motivation}

Domain adaptation through fine-tuning addresses several limitations of general-purpose NMT models: (1) limited exposure to domain-specific terminology and stylistic conventions, (2) suboptimal performance on specialized content due to distribution mismatch between pre-training and target domains, and (3) need for offline deployment in privacy-sensitive environments \cite{Koehn2017,Sennrich2016}. Our work demonstrates that carefully curated small-scale datasets (25k pairs) combined with memory-efficient training techniques can produce substantial quality improvements on consumer hardware, making domain adaptation accessible beyond large research institutions.

\section{Related Work}

\subsection{Multilingual Neural Machine Translation}

The transformer architecture \cite{Vaswani2017} enabled training of large-scale multilingual NMT systems. M2M-100 \cite{Fan2021} pioneered direct many-to-many translation without English pivoting across 100 languages. mBART \cite{Liu2020} demonstrated the effectiveness of denoising pre-training for multilingual sequence-to-sequence models. The NLLB project \cite{NLLB2022} extended this to 200+ languages through specialized data mining, iterative back-translation \cite{Sennrich2016back}, and careful balancing of high- and low-resource languages. NLLB models are available in multiple sizes (600M, 1.3B, 3.3B, 54B parameters), with distilled variants offering practical deployment tradeoffs \cite{Hinton2015}.

\subsection{Domain Adaptation for NMT}

Domain adaptation techniques for NMT include continued training on in-domain data \cite{Luong2015}, mixed fine-tuning with regularization \cite{Chu2017}, and instance weighting \cite{Wang2017}. Recent work on parameter-efficient fine-tuning through adapters \cite{Bapna2019} and LoRA \cite{Hu2022} enables domain adaptation with minimal memory overhead for decoder-only models, though application to encoder-decoder architectures remains less mature. Our work follows the continued training approach with conservative hyperparameters to balance adaptation and catastrophic forgetting \cite{French1999}.

\subsection{Parallel Corpus Construction}

OPUS \cite{OPUS2016} aggregates parallel corpora from diverse sources including legislative documents (Europarl \cite{Koehn2005}), subtitles (OpenSubtitles), literature (Books), and web-crawled content (ParaCrawl \cite{Banon2020}). Data quality significantly impacts fine-tuning effectiveness \cite{Khayrallah2018}. Prior work has shown that careful filtering (length ratios, language identification \cite{Joulin2017}, deduplication) improves downstream performance more than simply increasing corpus size \cite{Schwenk2019}. We apply aggressive multi-stage filtering to extract high-quality subsets optimized for literary translation.

\section{Deployment Context}

Our fine-tuned NLLB model is deployed within LinguaHaru \cite{LinguaHaruGitHub}, an existing open-source document translation system. LinguaHaru implements a four-stage pipeline: (1) extraction of text and layout from PDF/Office documents using PyMuPDF and python-docx, (2) segmentation into translation units with provenance tracking, (3) translation via pluggable backends (LLM APIs, local models, or NLLB), and (4) reconstruction preserving original formatting. The modular architecture allows transparent replacement of translation backends without modifying the document processing pipeline. Our contribution is the fine-tuned NLLB backend, while the format-preserving infrastructure is provided by the existing LinguaHaru codebase.

\section{Dataset Collection from OPUS}

We extract and curate a German--English parallel corpus from OPUS Books \cite{OPUS2016} for fine-tuning NLLB models. OPUS Books contains professionally translated literary works with high alignment quality, making it suitable for literary and technical domain adaptation.

\subsection{Extraction Pipeline}

We targeted the OPUS Books subcorpus for its curated literary translations with sentence-level alignments. Our extraction script (\texttt{scripts/prepare\_opus\_dataset.py}) implements an eight-stage filtering pipeline:

\begin{enumerate}
\item \textbf{Corpus Download and Initialization}: Download or locate OPUS Books parallel files for German--English from the OPUS API. The corpus is distributed in TMX (Translation Memory eXchange) format with pre-aligned sentence pairs.

\item \textbf{Encoding Validation}: Parse aligned sentence pairs and remove pairs with non-UTF8 or malformed encodings. This step filters approximately 2\% of raw pairs that contain encoding artifacts or corrupted characters.

\item \textbf{Length Filtering}: Apply preliminary length filters to retain pairs where both source and target are between 10 and 200 characters. This removes fragments (headlines, author names) and excessively long segments that may be mis-aligned paragraphs rather than sentences. The 10-200 character range captures typical literary sentence lengths while excluding outliers.

\item \textbf{Length Ratio Filtering}: Remove pairs with extreme length ratios (target/source outside [0.5, 2.0]). German and English have relatively similar sentence structures; extreme ratios typically indicate alignment errors or one-to-many sentence mappings.

\item \textbf{Language Identification}: Apply language detection using the \texttt{langdetect} library to confirm language assignment. Pairs where either source or target are detected as a different language (with confidence $>$ 0.9) are removed. This filters code-switched segments and remaining alignment errors.

\item \textbf{Deduplication}: Remove exact duplicate pairs using hash-based deduplication. Then apply fuzzy deduplication using normalized Levenshtein distance (threshold 0.95) to remove near-duplicates that differ only in punctuation or whitespace. Literary corpora often contain repeated boilerplate (copyright notices, chapter headings) that should not dominate the training distribution.

\item \textbf{Quality Scoring}: Compute a quality score for each pair based on: (a) character overlap ratio between source and target (penalizing untranslated content), (b) presence of unusual characters or markup artifacts, and (c) sentence-level perplexity from a pre-trained language model. Pairs below the 10th percentile quality score are discarded.

\item \textbf{Stratified Splitting}: Sort remaining pairs by length and perform stratified sampling to ensure balanced length distributions across splits. Randomly shuffle within strata and split into train (80\%), validation (10\%), and test (10\%). This ensures validation and test sets cover the full range of sentence lengths seen in training.
\end{enumerate}

The resulting dataset contains 25,000 high-quality sentence pairs after filtering from an initial pool of approximately 45,000 raw pairs. The scripts save JSONL files (one pair per line) and a single combined JSON used by the fine-tuning pipeline. Dataset splits and metadata are included in the repository's \texttt{data/} directory with MD5 checksums for reproducibility.

\subsection{Filtering rationale and statistics}

The chosen filters prioritize alignment quality over raw size, because noisy sentence pairs harm fine-tuning on relatively small datasets. After filtering, the corpus statistics were approximately:
We report the key dataset statistics in Table~\ref{tab:dataset}. These numbers come directly from our extraction and filtering pipeline and are included with the data artifacts in the repository.

\begin{table*}[t]
  \caption{OPUS-derived dataset statistics after filtering}
  \label{tab:dataset}
  \centering
  \small
  \begin{tabular}{lccc}
    \toprule
    \textbf{Split} & \textbf{\#Pairs} & \textbf{Avg. DE chars} & \textbf{Avg. EN chars} \\
    \midrule
    Train & 20,000 & 175 & 150 \\
    Validation & 2,500 & 172 & 149 \\
    Test & 2,500 & 168 & 154 \\
    \midrule
    Total & 25,000 & 171 & 151 \\
    \bottomrule
  \end{tabular}
\end{table*}

The table summarizes the exact split counts and average character lengths for source (German) and target (English). These aggregated statistics are also written as JSON metadata alongside the processed splits so they can be re-used by downstream training and analysis scripts.

\section{Fine-tuning NLLB}

This section describes the training recipe, tokenizer handling, and practical engineering decisions used to fine-tune an NLLB variant on the OPUS-derived 25k corpus.

\subsection{Base model and tokenizer}

We used \texttt{facebook/nllb-200-distilled-600M} as the base for fine-tuning because it offers a reasonable quality / memory tradeoff for consumer GPUs. Important tokenizer considerations included setting \texttt{tokenizer.src\_lang} and using \texttt{tokenizer.as\_target\_tokenizer()} when encoding targets to ensure correct language tokens and target-side tokenization behavior for NLLB's sentencepiece tokenizer.

Practical fixes implemented in the training script (\texttt{scripts/finetune\_nllb.py}) addressed common issues encountered during development:

\begin{itemize}
\item Explicitly set source/target language codes (e.g., \texttt{deu\_Latn} → \texttt{eng\_Latn}) to avoid tokenizer warnings and incorrect language prefixes.
\item Use small per-device batch sizes (1--4) and compensate with gradient accumulation when GPU memory is constrained.
\item Disable external TensorBoard reporting by default to avoid runtime dependency problems; logging to the console and local files were used instead.
\end{itemize}

\subsection{Training hyperparameters}

The main training hyperparameters used in the reported runs were:
We summarize the principal hyperparameters and practical settings used during fine-tuning in Table~\ref{tab:hyperparams}. These values are chosen to balance stability, training time, and memory constraints on consumer GPUs.

\begin{table}[t]
  \caption{Training hyperparameters and practical settings}
  \label{tab:hyperparams}
  \centering
  \footnotesize
  \begin{tabular}{p{0.42\columnwidth} p{0.5\columnwidth}}
    \toprule
    \textbf{Parameter} & \textbf{Value} \\
    \midrule
    Base model & facebook/nllb-200-distilled-600M \\
    Per-device batch size & 1--4 (GPU dependent) \\
    Gradient accumulation & 32 \\
    Effective batch size & 32 \\
    Learning rate & $2\times10^{-5}$ with linear warmup \\
    Warmup & 10\% of steps \\
    Max sequence length & 512 tokens \\
    Epochs & 3 (early stopping) \\
    Optimizer & AdamW, weight decay 0.01 \\
    Mixed precision & FP16 \\
    Tokenizer & SentencePiece with language codes \\
    Checkpointing & Best by validation BLEU \\
    Logging & Console + file \\
    \bottomrule
  \end{tabular}
\end{table}

These settings are also parameterized in the provided training script (\texttt{scripts/finetune\_nllb.py}) so that experiments can be re-run with minimal modification.

\subsection{Preprocessing and data pipeline}

Preprocessing uses the HuggingFace datasets API for efficient batched operations and memory-mapped dataset handling. The complete pipeline consists of:

\begin{enumerate}
\item \textbf{Data Loading}: Load JSONL splits produced by the OPUS preparation script using the \texttt{datasets.load\_dataset} function. Each example contains \texttt{source}, \texttt{target}, and optional metadata fields (quality score, length, original corpus ID).

\item \textbf{Tokenization}: Tokenize source sentences with \texttt{tokenizer(src\_lang='deu\_Latn')} and target sentences using \texttt{tokenizer.as\_target\_tokenizer()} context manager to ensure correct language-specific token prefixes and tokenization behavior. NLLB's sentencepiece tokenizer requires explicit language codes to generate proper language-identification tokens.

\item \textbf{Truncation and Padding}: Apply truncation to \texttt{max\_length=512} tokens, which accommodates 95\% of sentence pairs in our corpus. Longer sentences are truncated from the right, potentially losing tail context but preserving sentence beginnings which typically contain main clauses in German.

\item \textbf{Label Preparation}: Prepare decoder input IDs and labels by shifting target sequences appropriately. Labels for padding tokens are set to -100 to exclude them from loss computation following standard seq2seq training conventions.

\item \textbf{Batching}: Assemble DataCollatorForSeq2Seq to produce dynamically-padded batches. The collator pads sequences to the longest example in each batch rather than the global maximum, significantly reducing computational overhead for batches containing mostly short sentences.

\item \textbf{Caching}: Enable dataset caching (\texttt{.map()} with \texttt{cache\_file\_name}) to avoid re-tokenizing on subsequent training runs or when resuming from checkpoints.
\end{enumerate}

Checkpointing strategy uses validation BLEU and chrF++ scores computed every epoch. We save the best model by validation BLEU and the last checkpoint for potential training resumption. Early stopping triggers if validation BLEU does not improve for 2 consecutive evaluations, preventing overfitting on the relatively small training set.

\textbf{Evaluation Metrics}: We compute BLEU \cite{Papineni2002} using sacreBLEU with default tokenization (13a), chrF++ (character n-gram F-score with word bigrams), and TER (Translation Edit Rate). All metrics are computed on detokenized outputs to reflect real-world translation quality rather than tokenization artifacts.

\subsection{GPU and runtime considerations}

Fine-tuning transformer models on consumer GPUs requires careful memory management and optimization. Our implementation addresses several practical challenges:

\textbf{Memory Constraints}: The NLLB-600M model requires approximately 2.4GB for model weights in FP32. With optimizer states (AdamW maintains two momentum buffers per parameter), gradients, and activation memory, a naive implementation would exceed 10GB. Consumer GPUs like the RTX 3050 (4GB) or GTX 1660 (6GB) cannot accommodate this without optimization.

\textbf{Mixed-Precision Training}: We use automatic mixed precision (AMP) with FP16 activations and gradients, reducing memory footprint by nearly 50\%. Loss scaling prevents gradient underflow for small gradient values. This is enabled through HuggingFace Trainer's \texttt{fp16=True} flag, which internally uses PyTorch's native AMP implementation.

\textbf{Gradient Accumulation}: With per-device batch size limited to 1-2 on 4GB GPUs, we use gradient accumulation over 32 steps to simulate an effective batch size of 32. This trades training time for memory efficiency: gradients are accumulated across multiple forward-backward passes before a single optimizer step.

\textbf{Gradient Checkpointing}: Activation checkpointing (also called gradient checkpointing) trades computation for memory by recomputing activations during the backward pass rather than storing them. This reduces peak memory by 30-40\% at the cost of approximately 20\% slower training. Enable via \texttt{gradient\_checkpointing=True} in the model configuration.

\textbf{Training Time}: On an RTX 3050 (4GB), training for 3 epochs over 20k pairs with batch size 1 and gradient accumulation 32 requires approximately 6-8 hours. This includes validation evaluation every epoch. On higher-end consumer GPUs (RTX 3080, RTX 4070) with 10-12GB memory, batch size can increase to 4-8, reducing training time to 2-3 hours.

\textbf{Disk I/O Optimization}: We use memory-mapped datasets (HuggingFace \texttt{datasets} library default) to avoid loading the entire dataset into RAM. Checkpoint saving is configured to keep only the best model and last checkpoint, limiting disk usage to approximately 5GB.

These optimizations make fine-tuning feasible on consumer hardware while maintaining training stability and convergence. Full training runs on larger models (NLLB-1.3B, 3.3B) or substantially larger datasets will require cloud instances with 16GB+ GPUs or multi-GPU training.

\section{Experimental Results}

\subsection{Training Setup}

We fine-tuned \texttt{facebook/nllb-200-distilled-600M} on our curated 25k German--English corpus using the hyperparameters in Table~\ref{tab:hyperparams}. Training was performed on an NVIDIA RTX 3050 (4GB) with FP16 mixed precision, batch size 1, and gradient accumulation over 32 steps. The model converged after 3 epochs (1,875 total steps, approximately 6-8 hours), with best validation BLEU achieved at epoch 2. Peak GPU memory usage was 3.8GB (95\% utilization).

\subsection{Translation Quality Evaluation}

We evaluated translation quality on the held-out test set (2,500 pairs) using BLEU \cite{Papineni2002}, chrF++, and TER metrics computed with sacreBLEU. Table~\ref{tab:metrics} shows results with bootstrap confidence intervals (1,000 iterations). The fine-tuned model achieves statistically significant improvements (p $<$ 0.01, paired bootstrap test) across all metrics.

\begin{table}[h]
  \caption{Translation quality on held-out test set (2,500 pairs)}
  \label{tab:metrics}
  \centering
  \small
  \begin{tabular}{lrrr}
    \toprule
    \textbf{Model} & \textbf{BLEU} & \textbf{chrF++} & \textbf{TER} \\
    \midrule
    NLLB-200 Base & 32.4 & 58.7 & 48.2 \\
    Fine-tuned (25k) & 35.8 & 61.3 & 45.0 \\
    Improvement & +3.4 & +2.6 & -3.2 \\
    \midrule
    95\% CI (Bootstrap) & ±0.6 & ±0.4 & ±0.5 \\
    \bottomrule
  \end{tabular}
\end{table}

TER (Translation Edit Rate) is included as an additional complementary metric to BLEU and chrF++; lower TER indicates fewer edits required to match the reference. All reported numbers are computed with sacreBLEU/chrf/TER implementations from the \texttt{sacrebleu} package (version 2.3.1) used in the repository's evaluation script.

\textbf{Statistical Significance}: We performed bootstrap resampling (1,000 iterations) to compute 95\% confidence intervals. The improvements are statistically significant (p $<$ 0.01, paired bootstrap test), indicating that fine-tuning provides consistent quality gains beyond random variation.

\textbf{Metric Interpretation}: The +3.4 BLEU improvement is substantial for literary domain adaptation. For context, professional human translation typically scores 55-60 BLEU against a single reference, while base NLLB's 32.4 places it in the "usable but requires post-editing" range. Our fine-tuned 35.8 approaches "good first draft" quality for literary content.

\textbf{Length-Stratified Analysis}: We analyzed performance by sentence length:

\begin{table}[h]
  \caption{BLEU scores by sentence length category}
  \label{tab:length-bleu}
  \centering
  \small
  \begin{tabular}{lrrr}
    \toprule
    \textbf{Length (tokens)} & \textbf{Count} & \textbf{Base} & \textbf{Fine-tuned} \\
    \midrule
    Short (5-15) & 612 & 38.2 & 41.5 (+3.3) \\
    Medium (16-40) & 1,456 & 33.1 & 36.9 (+3.8) \\
    Long (41-100) & 432 & 26.7 & 30.2 (+3.5) \\
    \bottomrule
  \end{tabular}
\end{table}

Fine-tuning provides consistent improvements across all length ranges, with slightly larger gains on medium-length sentences (16-40 tokens) that represent typical literary sentence complexity.

\subsection{Deployment in LinguaHaru}

We integrated the fine-tuned model into LinguaHaru by configuring the model path in \texttt{system\_config.json}. The modular architecture required no code changes; switching between base and fine-tuned NLLB models involves only configuration updates. Translation speed on representative documents (RTX 3050, batch size 8) ranges from 13.9 to 17.1 pages/minute, enabling practical offline document translation.

\subsection{Error Analysis}

Manual evaluation of 100 randomly-sampled test translations revealed error distributions shown in Table~\ref{tab:errors}. Fine-tuning reduces total errors from 69 to 46, with largest improvements in awkward phrasing (-33\%) and mistranslations (-50\%). Common remaining issues include rare technical terminology not in the literary corpus, German compound nouns requiring cultural knowledge, and pronoun resolution across sentence boundaries. These limitations are inherent to sentence-level translation and could be addressed through document-level context modeling or terminology glossaries \cite{Dinu2019}.

\begin{table}[h]
  \caption{Error type frequency (N=100 test samples)}
  \label{tab:errors}
  \centering
  \small
  \begin{tabular}{lrr}
    \toprule
    \textbf{Error Type} & \textbf{Base} & \textbf{Fine-tuned} \\
    \midrule
    Rare terminology & 18 & 12 \\
    Awkward phrasing & 24 & 16 \\
    Gender/pronoun errors & 9 & 6 \\
    Omissions & 7 & 4 \\
    Mistranslations & 6 & 3 \\
    \midrule
    Total errors & 64 & 41 \\
    \bottomrule
  \end{tabular}
\end{table}
\begin{itemize}
\item \textbf{Academic Publishing}: Translation of research papers maintaining LaTeX formatting and citations
\item \textbf{Technical Documentation}: Multilingual software manuals preserving screenshots and code blocks
\item \textbf{Legal Documents}: Contract translation with strict formatting requirements
\item \textbf{Corporate Communications}: Internal reports and presentations requiring offline processing
\end{itemize}

Each domain benefits from the combination of format preservation, customizable translation quality through fine-tuning, and privacy-preserving offline operation.

\subsection{Future Directions}

Potential improvements include:

\begin{itemize}
\item Multi-language fine-tuning using adapter-based approaches for efficient parameter sharing
\item Integration of terminology glossaries directly into NLLB tokenization for consistent term translation
\item Quantization for reduced memory footprint (4-bit/8-bit) enabling deployment on mobile devices
\item Automatic domain detection for intelligent model selection based on document content
\item Support for document-level context in longer translations using memory-efficient attention mechanisms
\item Integration with translation memory systems for leveraging previous translations
\item Active learning pipelines for continuous model improvement from user corrections
\end{itemize}

\section{Conclusion}

This paper demonstrates practical fine-tuning of NLLB-200 for German--English translation using curated parallel corpora from OPUS Books. Our contributions include: (1) a systematic eight-stage dataset extraction methodology reducing 45k raw OPUS pairs to 25k high-quality examples through encoding validation, length filtering, language identification, deduplication, and quality scoring; (2) memory-efficient fine-tuning on 4GB consumer GPUs using FP16, gradient accumulation, and gradient checkpointing, achieving statistically significant improvements (+3.4 BLEU, p $<$ 0.01); and (3) practical deployment in LinguaHaru, demonstrating format-preserving document translation for PDF and Office formats.

The complete workflow, dataset preparation, training scripts, and evaluation code is released open-source with pinned dependencies for reproducibility. Our methodology is applicable to other language pairs and domains within the OPUS ecosystem. Future work includes expanding to additional language pairs, integrating terminology glossaries \cite{Dinu2019}, and exploring document-level context modeling to address remaining limitations in pronoun resolution and long-range dependencies.

\bibliographystyle{IEEEtran}

\end{document}